\pdfoutput=1

\documentclass[11pt]{article}

\usepackage[]{ACL2023}

\usepackage{times}
\usepackage{latexsym}
\usepackage{graphicx}
\usepackage[T1]{fontenc}

\usepackage[utf8]{inputenc}

\usepackage{microtype}

\usepackage{inconsolata}
\usepackage{multirow}
\usepackage{xcolor}
\usepackage{amssymb}
\usepackage{amsmath}
\usepackage{xtab}
\usepackage{tabularx}
\title{T-Rex: Text-assisted Retrosynthesis Prediction}

\author{Yifeng Liu$^{*\alpha}$, Hanwen Xu$^{*\beta}$, Tangqi Fang$^\alpha$, Haocheng Xi$^\alpha$, \\\textbf{Zixuan Liu$^\beta$, Sheng Zhang$^\gamma$,
Hoifung Poon$^\gamma$, Sheng Wang$^{\dagger\beta}$} \\
  $^\alpha$Institute for Interdisciplinary Information Sciences, Tsinghua University \\
  $^\beta$ Paul G. Allen School of Computer Science and Engineering, University of Washington\\
  $^\gamma$ Microsoft Research \\
  \texttt{\{liuyifen20, fangtq20, xihc20\}@mails.tsinghua.edu.cn}, \\
  \texttt{\{xuhw, zucksliu, swang\}@cs.washington.edu}, \\
  \texttt{\{zhang.sheng, hoifung\}@microsoft.com}
}
\begin{document}
\maketitle
\begin{abstract}
\renewcommand{\thefootnote}{}
\footnotetext{*: Co-first author. $\dagger$: Corresponding author}
As a fundamental task in computational chemistry, retrosynthesis prediction aims to identify a set of reactants to synthesize a target molecule. Existing template-free approaches only consider the graph structures of the target molecule, which often cannot generalize well to rare reaction types and large molecules. Here, we propose T-Rex, a text-assisted retrosynthesis prediction approach that exploits pre-trained text language models, such as ChatGPT, to assist the generation of reactants. T-Rex first exploits ChatGPT to generate a description for the target molecule and rank candidate reaction centers based both the description and the molecular graph. It then re-ranks these candidates by querying the descriptions for each reactants and examines which group of reactants can best synthesize the target molecule. We observed that T-Rex substantially outperformed graph-based state-of-the-art approaches on two datasets, indicating the effectiveness of considering text information. We further found that T-Rex outperformed the variant that only use ChatGPT-based description without the re-ranking step, demonstrate how our framework outperformed a straightforward integration of ChatGPT and graph information. Collectively, we show that text generated by pre-trained language models can substantially improve retrosynthesis prediction, opening up new avenues for exploiting ChatGPT to advance computational chemistry. And the codes can be found at \url{https://github.com/lauyikfung/T-Rex}.
\end{abstract}

\section{Introduction}
\begin{figure}[!t]
\centering
\includegraphics[width=\linewidth]{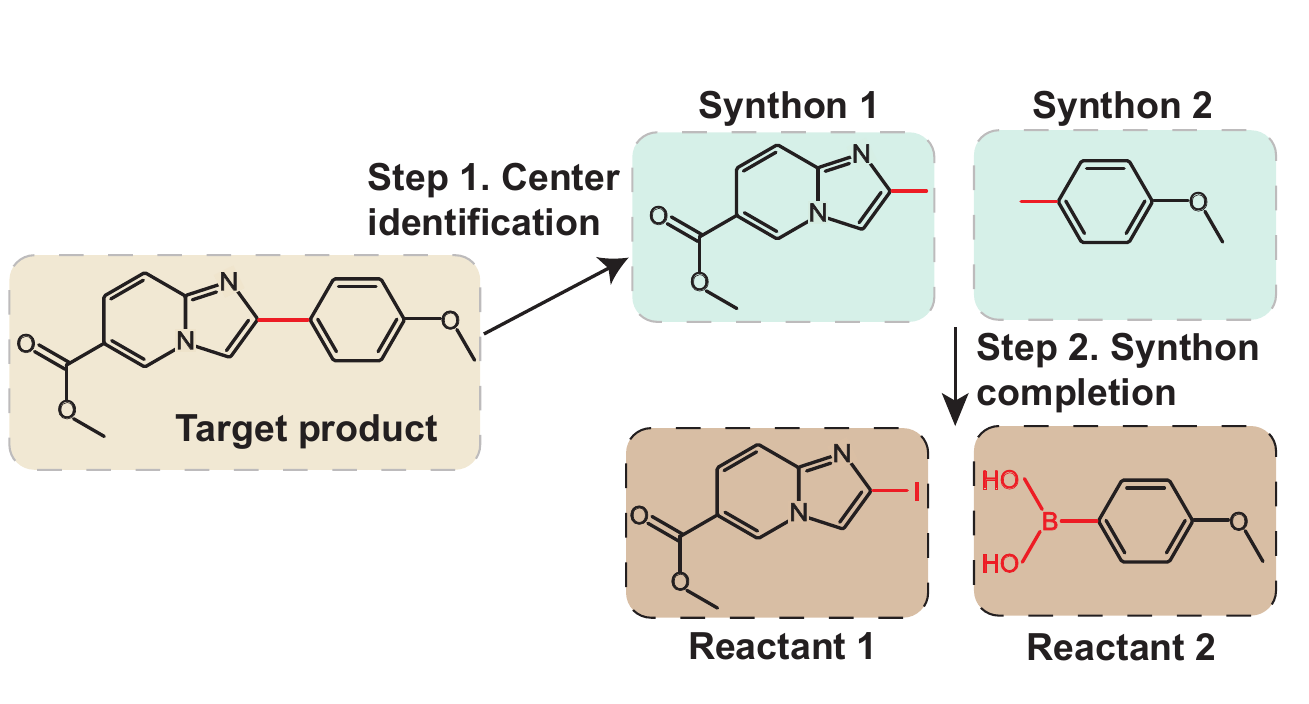}
\caption{\textbf{Illustration of the retrosynthesis prediction.} We formulate retrosynthesis prediction as a two-step approach. First, we identify the bond that splits the target molecule into two synthons. This step is formulated as a multi-class classification problem. Second, each synthon is used to generate a reactant. This step is formulated as a graph-to-graph generation problem.}
\label{intro}
\end{figure}

Retrosynthesis, which aims to find a few small reactants that can synthesize a given target product, is a fundamental problem in organic chemistry (\textbf{Fig.}~\ref{intro})~\cite{corey1988robert, szymkuc2016computer, segler2018planning, strieth2020machine, jiang2022artificial}. Retrosynthesis analysis is time-consuming and challenging even for experienced chemists due to the extremely large search space of all possible transformations. Recently, deep learning approaches have been proposed for retrosynthesis prediction~\cite{coley2018machine}. For example, the state-of-the-art retrosynthesis prediction method formulates this task as a graph generation problem, where a large graph of the target product is used to generate two small graphs of reactants. 

Existing retrosynthesis prediction approaches can be classified into template-based and template-free. Template-based approaches~\cite{hartenfeller2011collection,szymkuc2016computer, coley2017prediction, law2009route} match the subgraph patterns to templates, decompose the target molecule, and then use the indicated atomic modifications to obtain the reactants. Despite the substantial progress, template-based approaches are inefficient since the process of subgraph matching can yield hundreds of potential templates for a target molecule. As a result, template-free methods have become a promising alternative. Template-free methods utilize sequence-to-sequence techniques to generate target reactants using SMILES~\cite{SMILES} strings of the product~\cite{liu2017retrosynthetic, karpov2019transformer, tu2022permutation, sacha2021molecule, Dual-TF}. Recently, MEGAN~\cite{MEGAN} and G2Gs~\cite{G2Gs} have achieved the state-of-the-art retrosynthesis prediction results by transforming SMILES strings into graph structures and then exploiting a graph-to-graph generation framework.

Here, we propose a novel ranking and re-ranking approach T-Rex for retrosynthesis prediction. The key idea of T-Rex is to use large language models, including ChatGPT, to provide descriptions for the reaction procedure, and then integrate these descriptions with molecule graphs for retrosynthesis prediction. In particular, we first use ChatGPT to obtain the description of the target product based on its SMILES representation, and use this textual description and the molecular graph to rank candidate reaction centers. For each candidate reaction center, we then use ChatGPT to obtain the descriptions for both reactants. Finally, we use these description to re-rank the candidate reaction centers. After the candidate reaction center is identified, we exploit existing graph-to-graph approach for synthon completion.

We evaluated T-Rex on two large-scale retrosynthesis datasets USPTO-50k~\cite{USPTO-50k} and USPTO-MIT~\cite{USPTO-MIT}. We found that T-Rex substantially outperformed G2Gs model on both within-dataset cross-validation and cross-dataset prediction. We further demonstrated that T-Rex framework can be used to predict reaction type and these predicted reaction types can further boost the retrosynthesis prediction. Finally, we extended T-Rex to deconstruct a product to three reactants and showed that T-Rex can also achieve the best performance on this task.  Collectively, T-Rex utilized text data generated by language models to boost retrosynthesis prediction, opening up new avenues for using language models for computational chemistry. 

\section{Preliminaries}
\textbf{Retrosynthesis prediction} In this paper, a molecular graph is described as $G=(\mathbf{V},\mathbf{E})$ with $\mathbf{V}\in\{0,1\}^{n\times n\times b}$ and $\mathbf{E}\in\{0,1\}^{n\times d}$, where the numbers of atoms, bond types and the dimension of node features are $n,b,$ and $d$, respectively. $\mathbf{V}_{ijk} = 1$ if and only if atom node $i$ and atom node $j$ has bond type $k$. $\mathbf{E}$ is the matrix of the atom node features. 
\begin{figure*}[!t]
\centering
\includegraphics[width=\linewidth]{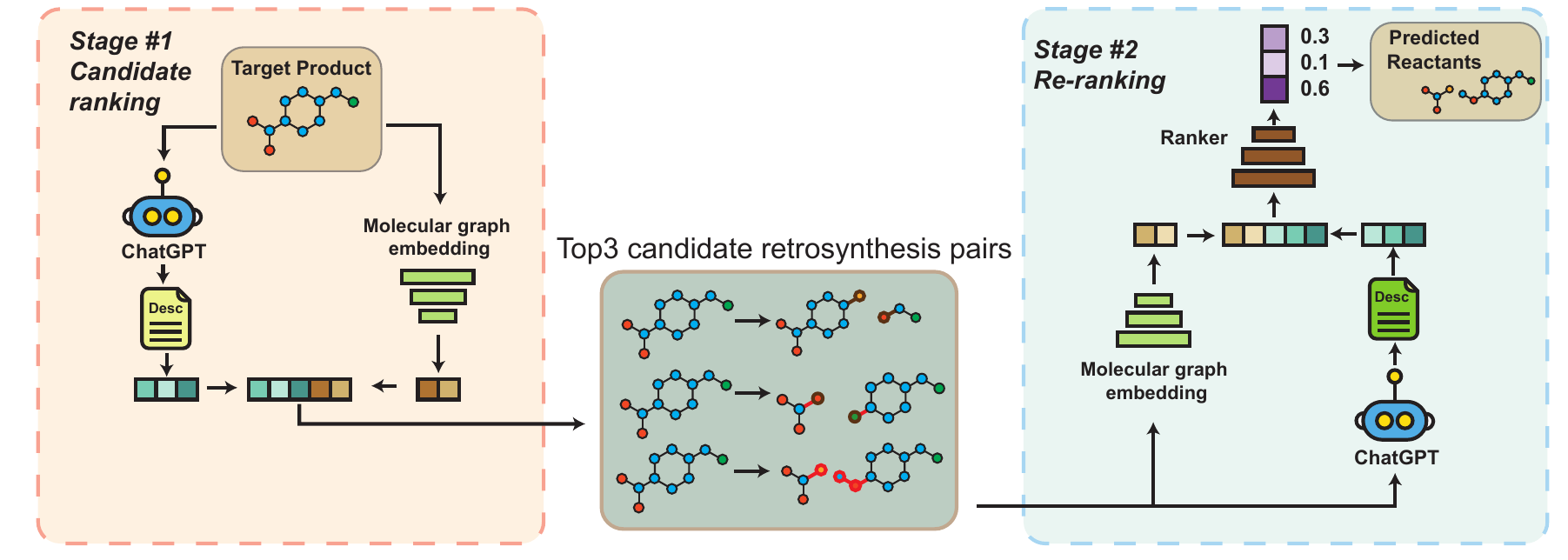}
\caption{\textbf{Diagram of T-Rex.} T-Rex is a two-stage approach. In the first stage, we use ChatGPT to generate a description for the target product. We then integrate this description and the molecular graph to obtain a few candidate reaction centers. In the second stage, we use ChatGPT to obtain a description for each synthon based on each candidate reaction center. The descriptions of two synthons are used together to re-rank the candidate reaction centers.}
\label{flowchart}
\end{figure*}
A retrosynthesis pair $R$ is represented by a pair of molecular graphs $(G_p, G_r)$. Here $G_p=\{G_{p_c}\}_{c=1}^{C_p}$ are the products and $G_r=\{G_{r_c}\}_{c=1}^{C_r}$ are the reactants. In this paper, we focus on the single product case, therefore $C_p=1$ and we simplify the $R$ as $(G_p, \{G_{r_c}\}_{c=1}^{C_r})$.\\
\textbf{Reaction center identification.} We follow G2Gs~\cite{G2Gs} by defining reaction center as an atom pair $(i, j)$ so that there is no bond between $i^{th}$ and $j^{th}$ nodes in $G_r$ but there is a bond between $i^{th}$ and $j^{th}$ nodes in $G_p$. The task of reaction center identification is to find the reaction centers to get the synthons $\{S_c\}_{c=1}^{C_r}$ that are subgraphs extracted from $G_p$ by simply breaking down the according bonds. Since the reaction centers could be on the rings, there can be only one synthon(i.e., one reactants) for a single product. And the synthons may not be a valid molecule, which can be transformed into target reactants by the reactant generation task.\\
\textbf{Reactant generation.} For a retrosynthesis pair $R=(G_p, \{G_{r_c}\}_{c=1}^{C_r})$, denote $\{S_c\}_{c=1}^{C_r}$ as the according synthons of $G$. The reactant generation task is to predict the reactants $\{G_{r_c}\}_{c=1}^{C_r}$ by these synthons. Since in different retrosynthesis pairs, the same synthon may correspond to different reactants, the information of products should also be considered, which is represented by a latent vector $z$ in G2Gs model.\\
\textbf{Molecular graph feature learning.}\label{L-layer-GCN} To encode the overall characteristics of the graph in both tasks, the G2Gs model introduced an L-layer Relational Graph Convolutional Networks(R-GCN)~\cite{R-GCN} to get the node embeddings and graph-level embeddings. Let $s\in \mathbb{R}$ be the embedding dimension and $\mathbf{h}^l\in \mathbb{R}^{n\times s}$ be the hidden state at the $l^{th}$ layer computed by R-GCN, where $\mathbf{h}^0=\mathbf{E}$. For each layer, the R-GCN compute the hidden state by:
\begin{align}
    \mathbf{h}^l=\text{Ag}(\text{ReLU}(\mathbf{D}_i\mathbf{h}^{l-1}\mathbf{W}_i^l|i\in(1,\cdots,b))
\end{align}
where $\mathbf{D}_i=\mathbf{V}_{[:,:,i]}+I$ is the adjacency matrix for the $i^{th}$ bond type with self-loop and $W_i^l$ is the trainable parameters. Ag$(\cdot)$ is an aggregation function selected from summation, averaging and concatenation. The overall graph embedding $\mathbf{h}_G$ is calculated by summation of $\mathbf{h}^L$, i.e., $\mathbf{h}_G=\text{Sum}(\mathbf{h}^L)$.

\section{Methodology}
Our proposed framework T-Rex is a novel text-assisted retrosynthesis prediction approach. The core idea of T-Rex is to introduce the power of modern large-scale language models to enhance the current molecular structure-based representations. Following previous approaches, we formulated the retrosynthesis task as the reaction center identification and synthon completion. Notably, reaction center identification is a prerequisite for robust synthon completion and can be a bottleneck for retrosynthesis. Therefore, our approach investigated how textual descriptions could correct the mismatched reaction centers. We introduced a two-stage center identification strategy, based on ranking and re-ranking of candidate reaction centers with their textual descriptions. In the first stage, we integrate the text information into a graph neural network to roughly identify a few candidate reaction centers. In the second stage, we exploit a molecule-text translation framework to train a re-ranking model. The re-ranking model is capable of recognizing the correct reaction centers from a list of reaction centers identified by the first-stage model. This end-to-end two-stage framework finally enables accurate and robust molecule retrosynthesis, as illustrated in \textbf{Fig.}~\ref{flowchart}.
\subsection{Candidate ranking using ChatGPT}
We first denoted a retrosynthesis pair as $R=(G_p, \{G_{r_c}\}_{c=1}^{C_r})$. In reaction center identification, we used a binary matrix $\mathbf{B}\in\{0,1\}^{n\times n}$ as the labels for reaction centers, where $\mathbf{B}_{ij}=1$ recognized the chemical bond between $i^{th}$ node and $j^{th}$ node as the reaction center. Therefore, the center identification task was formulated as a binary prediction task. The input of this task is the structure information of the product molecules, e.g., the SMILES sequence, and the output is the probability of each bond being identified as the reaction center.

To obtain a high-quality molecule structure representation, we integrate the textual information and graphic information together. We first get the text description of the products by ChatGPT, which generates descriptions based on the IUPAC name of the product. The IUPAC name represents a unique representation of a chemical structure, based on a set of mapping rules between structures and language phrases, and is characterized in a more natural language-like format as compared to SMILES. Therefore, the IUPAC name can serve as a bridge between the chemical molecules and large-scale language models. Here we use the PubChemPy~\cite{PubChemPy} package to generate IUPAC name for each molecule in our dataset. We then generate the text description by the following prompt:
\\
\\
\fbox{%
  \parbox{\columnwidth}{
  \textit{
Please delineate the structural features, functional aspects, and applicable implementations of the molecule \{\{ NAME \}\}, commencing with the introduction:"The molecule is \{\{ NAME \}\}". Reasoning the most plausible type for synthesizing this molecule in the final step, and offer a rationale for your choice.}}%
}\\
\\
where the \{\{ NAME \}\} item is replaced by the IUPAC name. For generated text description, we trained a BERT model to get the textual embedding $\mathbf{H}_t$. Then we also utilize a L-layer R-GCN introduced in \textbf{Sec.}~\ref{L-layer-GCN} for computing graphic embedding vector $\mathbf{H}_g^L$:
\begin{align}
\mathbf{h}^L=\text{R-GCN}_1(G_p), \mathbf{H}_g^L=\text{Sum}(\mathbf{h}^L).
\end{align}
After that, we obtain the embedding for each edge by cross concatenation:
\begin{align}
\mathbf{e}_{ij}=\mathbf{h}_i^L||\mathbf{h}_j^L||\mathbf{V}_{ij}||\mathbf{H}_g^L||\mathbf{H}_t
\end{align}
where $||$ stands for vector concatenation, $\mathbf{h}_i^L\in R^s$ denotes the $i^{th}$ row of $h^L$ representing the features of $i^{th}$ node and $\mathbf{V}_{ij}=\mathbf{V}_{[i,j,:]}\in\{0,1\}^b$ is the edge type between $i^{th}$ node and $j^{th}$ node. When accounting for the scenario with reaction type given, we can incorporate this information by just concatenating its embedding with $\mathbf{e}_{ij}$. The final probability of the chemical bond being the reaction center is computed by:
\begin{align}
\mathbf{r}_{ij}=\sigma(\text{FF}_1(\mathbf{e}_{ij})).
\end{align}
Here $\text{FF}_1$ denotes a feedforward network and $\sigma$ is the Sigmoid activation function. Subsequently, we decided the reaction center bond to be the one that connects the $i^{th}$ node and $j^{th}$ node with the highest $\mathbf{e}_{ij}$ value. The model was trained using the cross entropy loss\:
\begin{align}
\mathcal{L}_1=-\sum_{x\in X}&\sum_{i\ne j}\lambda \mathbf{B}_{ij}\log(\mathbf{r}_{ij})\nonumber\\
&+(1-\mathbf{B}_{ij})\log(1-\mathbf{r}_{ij}),
\end{align}
where X is the training dataset and $\lambda\in[1,+\infty)$ is a weighting hyper-parameter employed to balance the extremely small logits of $\mathbf{r}_{ij}$. 

After center identification, we then cut off the chemical bond of reaction centers to get the synthons where each synthon is a connected subgraph. Then we generate the reactants for each synthon using a synthon completion model. We simply utilize the same structure in \cite{G2Gs} to predict the candidate reactant pairs by beam search. We collected a list of most possible candidate pairs to train the model in re-ranking stage.

\subsection{Re-ranking using ChatGPT}
In the previous stage, we collected a lot of retrosynthesis pairs. However, it is notable that the generated reactants may mismatch the ground truth due to the limited capabilities GNN-based molecule representations. To determine the most possible reactants, we proposed the re-ranking stage by utilizing the capability of ChatGPT. Given a set of retrosynthesis pairs $S=\{R_i\}_{i=1}^k$ for the product $x$, where $k$ is the beam size in the synthon completion and $R_i=(G_p,\{G_{r_c}\}_{c=1}^{C_{r_i}})$ is an individual retrosynthesis pair sharing the same product with other pairs in the set. The re-ranking model is a classifier which identified the candidate reactants from several candidates which have been predicted as the retrosynthesis path with a high probability. The input of the re-ranking clasifier is the same as the first stage model, i.e., the concatenation of text embeddings and molecular graph embeddings. The textual description $T^i$ of the reactant candidate $R_i$ was generated by ChatGPT using a different prompt:\\
\\
\fbox{%
  \parbox{\columnwidth}{%
\textit{Please delineate the structural features, functional aspects, and applicable implementations of the molecules \{\{ NAME \}\} and possible reactants \{\{ REACTANT1 \}\} and \{\{ REACTANT2 \}\} to synthesize it. You should generate the descriptions respectively as above example. These descriptions are linked by " [SEP] ", and each commences with the introduction:"The molecule is ...".}
  }%
}\\
\\
where the \{\{ NAME \}\}, \{\{ REACTANT1 \}\} and \{\{ REACTANT2 \}\} are the IUPAC names of the product and the two reactants. In the case when there only exists one reactant, we repeat the IUPAC name twice to make it compatible with our paradigm. Then we used a BERT model as the encoder of the textual descriptions:
\begin{align}
    \mathbf{H}_t^i=\text{BERT}_2(T^i).
\end{align}
To embed the molecular graph information, we also utilized the R-GCN structure as the encoder:
\begin{align}
\mathbf{h}^{L,i}=\text{R-GCN}_2(\{G_{r_c}\}_{c=1}^{C_{r_i}}), \mathbf{H}_g^i=\text{Sum}(\mathbf{h}^{L,i}).
\end{align}
Here we exclude the graph of product since its information has been already extracted in the previous stage. Afterwards, the embedding for the retrosynthesis pair is obtained by concatenating the textual and graphic embedding vectors:
\begin{align}
\mathbf{H}^i=\mathbf{H}_g^i||\mathbf{H}_t^i.
\end{align}
Finally, we get a probability score for $R_i$ being recognized by the classifier:
\begin{align}
a^i=\text{Softmax}(\text{FF}_2(\mathbf{H}^i)),
\end{align}
where $\text{FF}_2$ is another feedforward network and $a^i=[a_-^i,a_+^i]$ is the computed logits from the classifier. The confidence score, i.e., the probability for each candidate is then denoted as $a_+^i$.

In our implementation, we selected the top 3 reactant candidate from the first stage model as the inputs to the re-ranking model. And we label the candidates matching the ground truth as positive examples and others as negative examples. Notably, we observed a significant data distribution imbalance, i.e., two-thirds of instances have negative labels among the generated candidates. To augment the number of positive samples in the training set of the re-ranking model, we additionally add the ground truth reactants with positive labels. Then we train the re-ranking model as a binary classification task. The model is trained using cross entropy loss for the classification task. Additionally, we incorporate a contrastive learning loss term to optimize the molecule representations. Finally, the total loss is computed as follows:
\begin{align}
\mathcal{L}_2=\sum_{S\in \mathbb{S}}\sum_{i}\text{CELoss}(a_S^i,label_S^i)\nonumber\\
-\alpha\sum_{j\ne i}\log\frac{\exp(\text{sim}(\mathbf{H}_t^i, \mathbf{H}_t^j)}{\sum_m\exp(\text{sim}(\mathbf{H}_t^i,\mathbf{H}_t^m))},
\end{align}
where $\text{sim}(\mathbf{u}, \mathbf{v})=\frac{\mathbf{u}^T\mathbf{v}}{||\mathbf{u}||\cdot||\mathbf{v}||}$ is the cosine similarity function and $label_S^i$ is the label for the $i^{th}$ retrosynthesis pair in set $S$. Furthermore, the two losses are appropriately balanced via the utilization of a parameter denoted as $\alpha$.
\begin{figure}[!t]
\centering
\includegraphics[width=\linewidth]{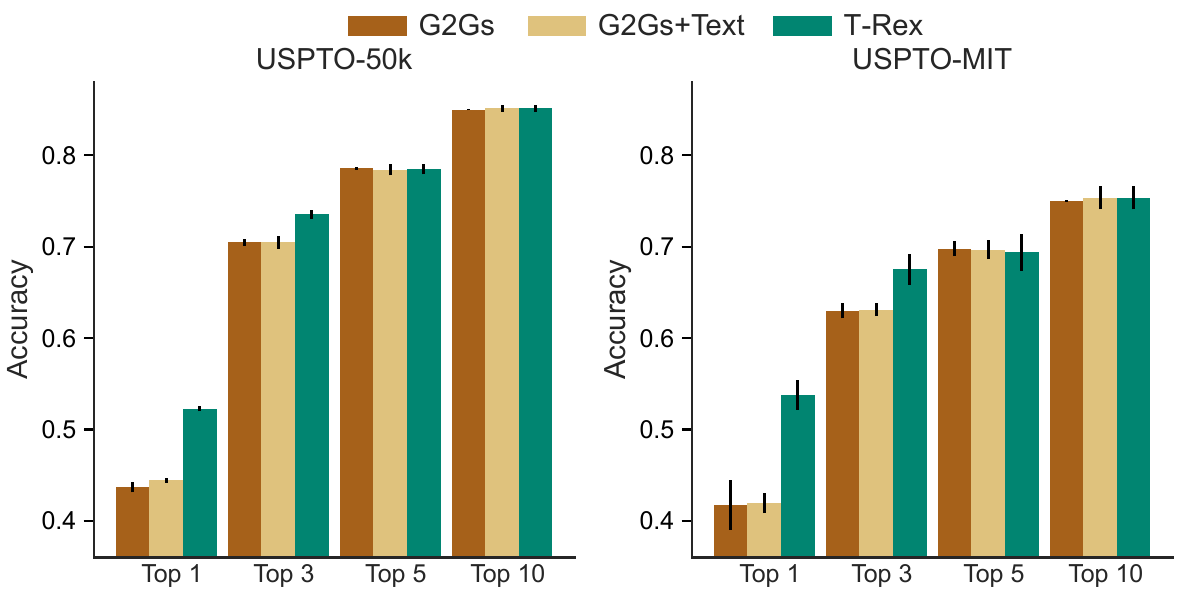}
\caption{\textbf{Performance on with-dataset cross-validation.} Top-k exact match accuracy on USPTO-50k and filtered USPTO-MIT datasets when reaction class is not given.}
\label{main_result}
\end{figure}
During the inference phase, we compute the probability for each reactant pair among the top $k$ generated candidates. In the evaluation, we can calculated the top $k'$ exact match accuracy by considering the reactant pairs with the largest $k'$ scores. In our implementation, we evaluate top $k'=1, 3, 5$ accuracy based on top $k=3, 5, 10$ candidates, respectively.
\section{Experimental setting}
We evaluated our method and comparison approaches on two widely-used retrosynthesis prediction datasets: USPTO-50k~\cite{USPTO-50k} and the filtered USPTO-MIT~\cite{USPTO-MIT}. All the reactions in both datasets are excluded in USPTO-MIT. The USPTO-50k dataset contains 50,016 reactions with 10 reaction types. The USPTO-MIT dataset does not contain the information for reaction type. In USPTO-MIT, we only considered reactions that have one product and two reactants. This results in 62,212 reactions. We followed previous work~\cite{G2Gs} to split the retrosynthesis pairs into 80\% training, 10\% validation and 10\% test using cross-validation.
\begin{figure*}[!t]
\centering
\includegraphics[width=\linewidth]{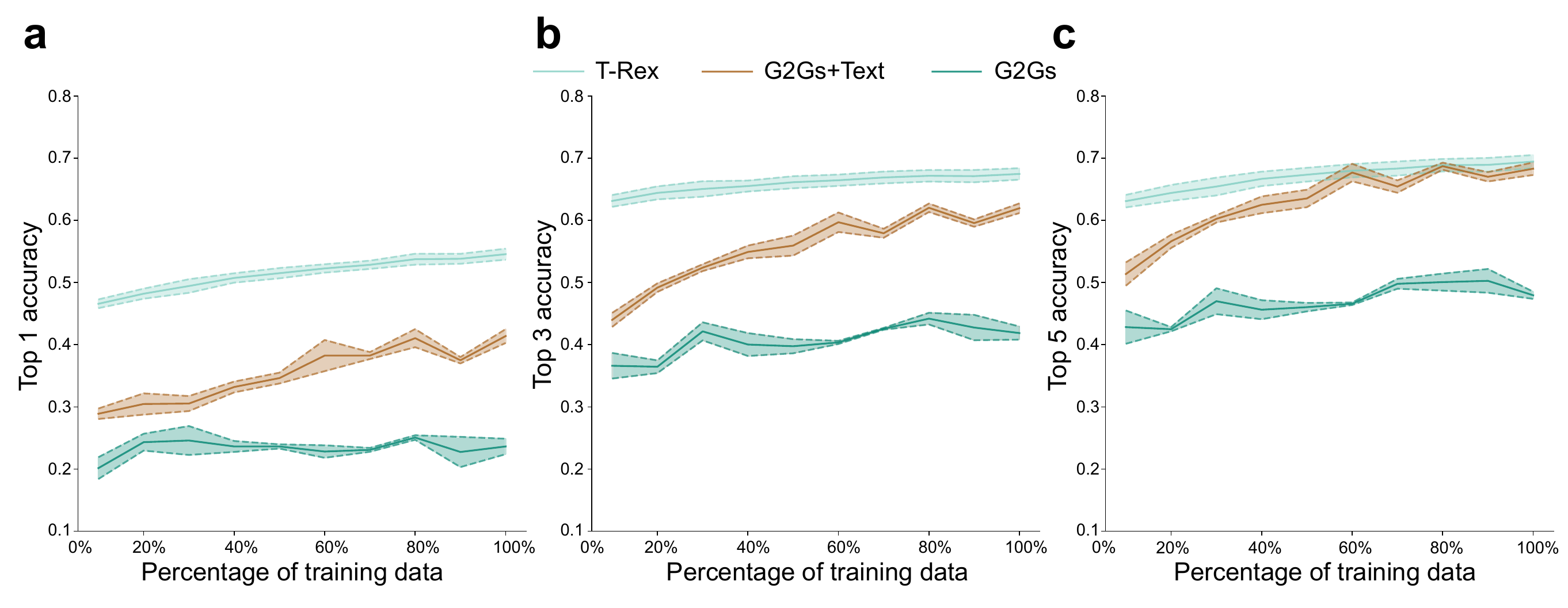}
\caption{\textbf{Comparison on cross-dataset retrosynthesis prediction.} Top 1, 3 and 5 exact match accuracy for step experiments of G2Gs, G2Gs+Text and our T-Rex model w.r.t. the percentage of the proportion of training set in filtered USPTO-MIT added for training. }
\label{step_exp}
\end{figure*}
We compared our method to three other template-free approaches. Particularly, \textbf{Seq2Seq2}~\cite{liu2017retrosynthetic} and \textbf{Transformer}~\cite{karpov2019transformer} are two template-free end-to-end frameworks directly generate SMILES strings of reactants from product SMILES strings. \textbf{G2Gs}~\cite{G2Gs} is one of the state-of-the-art template-free approaches which splits the prediction process into center identification and reactant generation. We obtained the results of G2Gs model from \url{https://torchdrug.ai/docs/benchmark/retrosynthesis.html}, which is slightly different from those in~\cite{G2Gs}. Following G2Gs~\cite{G2Gs}, we use the top-k exact match accuracy as our evaluation metrics where k is chosen from 1, 3, 5, and 10. Moreover, we also compute the center identification accuracy to examine the benefit of using text embeddings.

We used RDKit~\cite{RDKit} to preprocess SMILES strings. For generating text information, we used the Transformers library to generate the text information by MolT5~\cite{MolT5} and to get the text embedding by BERT-small~\cite{Bert-small1, Bert-small2} and PubMedBERT~\cite{pubmedbert}. And we also use the API for the gpt-3.5-turbo-0301 model provided by OpenAI to generate more detailed descriptions of the products with no penalty for frequency and presence and the temperature set as 0.7 and maximal token number as 512. We implement the R-GCN with 4 layers of the same embedding size of 512. And the graph embedding sizes for center identification and synthon completion tasks are set to 1536. We set $\lambda$ as 20 and the beam size as 10 in the candidate generation stage while the $\alpha$ is set to 0.2 in the re-ranking stage. We optimize our model with Adam~\cite{Adam}. The training hyperparameters are set as follows. We set the batch size of center identification batch size to 32 and the batch size of synthon completion to 128, and the batch size of ranking stage as 8. The learning rates for two stages are set to 1e-3 and 1e-5, respectively, and the warm-up ratio to 0.1. The weight decay is set to 0.01. We trained center identification model, synthon completion model and re-ranking model for 50, 10 and 10 epochs, respectively. We set beam size to 10 in beam search of retrosynthesis. All experiments are carried out on a single NVIDIA RTX A4000.

\section{Experimental results}
\label{result-section}
\subsection{Subtantial improvement on retrosynthesis prediction}
We first evaluated the retrosyntheis prediction performance using T-Rex. We calculated the top-k accuracy of the proposed approach on USPTO-50k and filtered USPTO-MIT datasets in both reaction class known and reaction class unknown settings. The results of T-Rex, G2Gs and G2Gs+text models when reaction class was not given is summarized in \textbf{Fig.}~\ref{main_result} and \textbf{Appendix.}~\ref{sec:appendixA}. We first noticed that G2Gs+Text outperformed the state-of-the-art template-free model G2Gs on Top 1 and Top 3 accuracy and achieved comparable performance on Top 5 and Top 10 accuracy on both datasets. This result demonstrates the effectiveness of incorporating text description generated by ChatGPT into the graph-to-graph retrosynthesis model. Furthermore, our T-Rex model exhibits  
 superior performance on top-1 and top-3 accuracy on two datasets compared to the G2Gs+Text model, highlighting the benefits of the re-ranking stage in the T-Rex model.

\subsection{Cross-dataset prediction}
We next examine the generalizability of our T-Rex model by evaluating the cross-dataset retrosythesis prediction. Specifically, we trained a model on USPTO-50k to predict the retrosythesis on USPTO-MIT \textbf{Table}~\ref{tab:simple_transfer}. Since USPTO-MIT and USPTO-50k have disjoint reaction types, these two datasets present substantially different distributions. We found that T-Rex achieved the best results on Top 1 and Top 5 in this cross-dataset setting. On the reverse side, our method also obtained the best performance on Top 1 and Top 5 accuracy, reflecting the good generalizability of our T-REX model.

Next, we conducted a series of incremental experiments by training on the entire set of USPTO-50k and part of USPTO-MIT, and testing on the remaining USPTO-MIT. We summarized the performance in \textbf{Fig.}~\ref{step_exp}. T-Rex consistently outperformed other approaches with regard to different proportions of USPTO-MIT training set. This further confirms the strong generalizability of T-Rex and the benefit of using text for retrosynthesis prediction.

Finally, since USPTO-MIT does not have reaction type information, we exploited T-Rex to predict the reaction type. Specifically, we used USPTO-50k to train a reaction type classifier and then used this classifier to predict reaction type for each target product in a multi-class classification setting. We then incorporated this reaction type into our framework. We found that this can further improve the retrosynthesis prediction performance (\textbf{Appendix} \ref{sec:appendixB}), again demonstrating the ability of T-Rex to jointly learn from heterogeneous datasets.

\subsection{Ablation studies for the re-ranking stage}
After observing substantial improvement of T-Rex over the G2Gs+Text baseline, we then conducted ablation studies for different components in the re-ranking stage. In particular, we assess the variants without text embedding (denoted as T-Rex w/o text), without graph embedding (denoted as T-Rex w/o graph), without contrastive learning loss on USPTO-50k (denoted as T-Rex w/o c.l.l.) or with the texts generated by MolT5 (denoted as T-Rex(MolT5)). The results of top 1, top 3 and top 5 exact match accuracy are shown in \textbf{Table}~\ref{tab:ticl}.
\begin{table}[h]
\centering
\begin{tabular}{lllll}
\hline
\multicolumn{1}{c}{\multirow{2}{*}{Models}} & \multicolumn{4}{c}{Accuracy}                                             \\ \cline{2-5} 
\multicolumn{1}{c}{}                        & Top-1             & Top-3             & Top-5             & Top-10                                \\ \hline
\multicolumn{5}{c}{USPTO-50k$\rightarrow$USPTO-MIT}                                                                           \\ \hline
G2Gs                   & 0.171          & 0.292          & 0.338          & \multicolumn{1}{c}{\textbf{0.396}} \\
G2Gs+Text              & 0.166          & 0.306          & \textbf{0.347} & \multicolumn{1}{c}{0.391}          \\
T-Rex                  & \textbf{0.204} & \textbf{0.307} & 0.329          & 0.391              \\ \hline
\multicolumn{5}{c}{USPTO-MIT$\rightarrow$USPTO-50k}                                                                           \\ \hline
G2Gs                  & 0.233             & 0.424             & \textbf{0.479}             & \textbf{0.548}                                 \\
G2Gs+Text             & 0.199             & 0.383             & 0.442             & 0.499                                 \\
T-Rex                 & \textbf{0.300} & \textbf{0.431} & 0.456 & 0.499                        \\ \hline
\end{tabular}
\caption{Results of the cross-dataset prediction.}
\label{tab:simple_transfer}
\end{table}
\begin{table}[h]
\centering
\begin{tabular}{lllll}
\hline
\multicolumn{1}{c}{\multirow{2}{*}{USPTO-50k}} & \multicolumn{3}{c}{Accuracy}                         \\ \cline{2-4} 
\multicolumn{1}{c}{}                         & Top-1             & Top-3             & Top-5             \\ \hline 
T-Rex w/o text                                & 0.453            & 0.633       & 0.647  \\ 
T-Rex w/o graph                                & 0.434            & 0.677       & 0.707  \\
T-Rex w/o c.l.l.                                 & 0.491        & 0.705          & 0.740          \\
T-Rex(MolT5)                                 & 0.509        & 0.712          & 0.755          \\\hline
T-Rex                                     & \textbf{0.523} & \textbf{0.735} & \textbf{0.785}  \\ \hline
\end{tabular}
\caption{Results of the ablation study for different components in the re-ranking stage.}
\label{tab:ticl}
\end{table}
\begin{figure*}[!t]
\centering
\includegraphics[width=\linewidth]{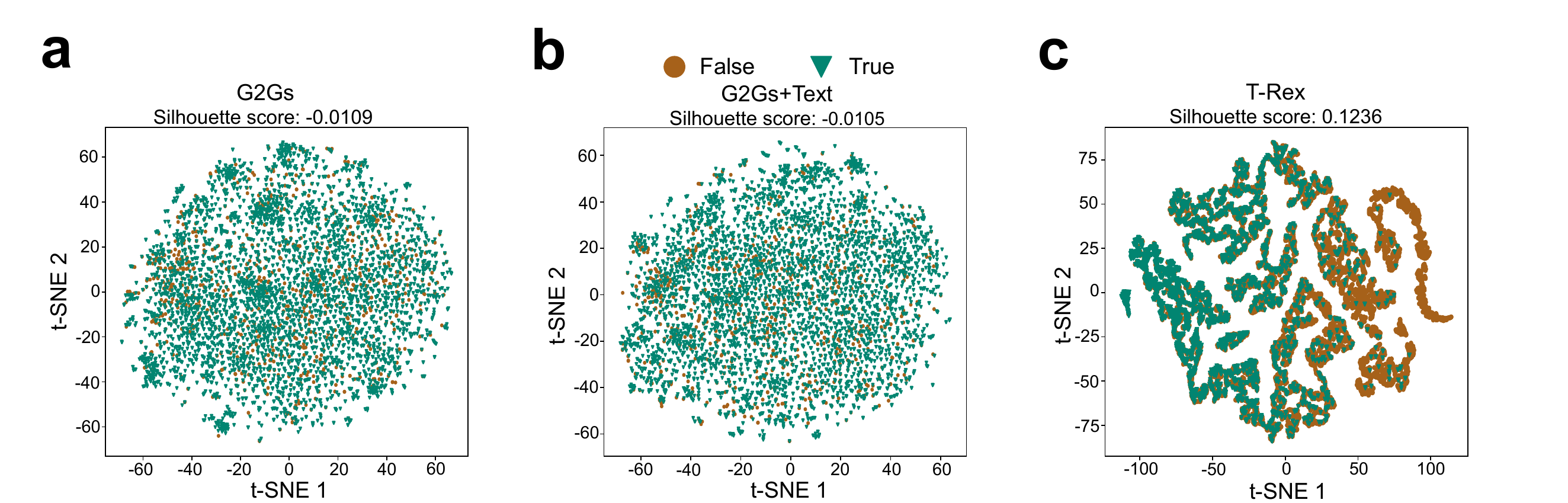}
\caption{\textbf{Visualization of the embedding space of three models.} G2Gs and G2Gs+Text show the embeddings from the reaction center identification stage. T-Rex shows the embeddings from the re-ranking stage, which are unavailable for the other two methods.}
\label{tsne}
\end{figure*}
We found that all these four components are important because the original T-Rex outperformed all these four variants. Moreover, the inclusion of text information contributes to performance improvements most substantially, as demonstrated by large drops in Top 3 and Top 5 accuracy by the variant that does not use the text information, necessitating the incorporation of text information into our framework. Moreover, the variant that uses MolT5 instead of ChatGPT to generate textual description performs substantially worse than T-Rex, indicating the importance of using high-quality text in our framework, further supporting the benefits of using text information for retrosynthesis prediction.
\subsection{Transferring Test on Predicted Reaction Type}
We also tried to improve the performance of our T-Rex model in the transferring test with the help of reaction type. Specifically, we started by training a reaction-type classifier on the USPTO-50k dataset. We then employed this classifier to predict the reaction types within the filtered USPTO-MIT dataset, even in cases where certain reactions cannot be definitively classified under any specific reaction type.

We have explored various potential classifier architectures, encompassing different combinations of graph features solely pertaining to the product (denoted as g.p.), graphic and textual information associated with the product (denoted as g.p.+t.p.), as well as graphic features of the reactants and product in conjunction with textual information specific to the product (denoted as g.p.+t.p.+g.r.). To make the experiments comparable, we only gave the reaction type information in the re-ranking stage. The top 1, 3, and 5 exact match accuracies are listed in \textbf{Table}~\ref{tab:MIT-type}.
\label{sec:appendixB}
\begin{table}[]
\centering
\begin{tabular}{llll}
\hline
\multicolumn{1}{c}{\multirow{2}{*}{USPTO-MIT}} & \multicolumn{3}{c}{Accuracy \%}                                             \\ \cline{2-4}
\multicolumn{1}{c}{}                       & Top-1             & Top-3             & Top-5             \\ \hline                                                     
g.p.                                        & 0.568          & 0.701          & 0.719         \\
g.p.+t.p.                                   & 0.568          & 0.700          & 0.718  \\
g.p.+t.p.+g.r.                              & \textbf{0.574} & \textbf{0.706} & \textbf{0.722}  \\ \hline
T-Rex(no reaction type)         & 0.523 & 0.735 & 0.785  \\ \hline
\end{tabular}
\caption{Results of the models trained on filtered USPTO-MIT dataset with the predicted reaction types.}
\label{tab:MIT-type}
\end{table}
We can see that with the help of predicted reaction types, the top 1 exact match accuracy can improve much, demonstrating its potential to incorporate information of new resources.
\subsection{Selection of Text Embedding Model}
In our experiment setting, we used two types of BERT: BERT-small and PubMedBERT. And we also did the experiments with only one types of BERT for USPTO-50k, and the results are presented in Table \ref{tab:bert}. We can notice that the selection of different types of embedding models may also contribute to the performance of the retrosynthesis prediction task. 
\begin{table}[h]
\centering
\begin{tabular}{llll}
\hline
\multicolumn{1}{c}{\multirow{2}{*}{USPTO-50k}} & \multicolumn{3}{c}{Accuracy \%}                                             \\ \cline{2-4}
\multicolumn{1}{c}{}                       & Top-1             & Top-3             & Top-5             \\ \hline                                                     
BERT-small-only                                       & 0.502          & 0.712          & 0.753         \\
PubMedBERT-only                                   & 0.509          & 0.713          & 0.756  \\
\multicolumn{1}{l}{\begin{tabular}[c]{@{}l@{}}BERT-small + \\ PubMedBERT (\%)\end{tabular}}        & \textbf{0.523} & \textbf{0.735} & \textbf{0.785}  \\ \hline
\end{tabular}
\caption{Results of the models trained on USPTO-50k for different BERT type combination.}
\label{tab:bert}
\end{table}

\subsection{Deep Investigation on Re-ranking Stage}
In Figure \ref{main_result}, we noticed T-Rex posed a great improvement than G2Gs on top-1 accuracy, but a smaller increase in top-3, 5 and 10. Therefore, we made deep investigation on the re-ranking stage in two aspects. Firstly, the increase in the top-1 accuracy results from the re-ranking stage. We have recorded the accuracies before and after re-ranking for each set of retrosynthesis pairs. For ground truth retrosynthesis pairs ranking first in the candidate ranking stage, 87.3\% of them remain top-1 by the ranker in the re-ranking stage. And for those ranking 2 or 3 in the first stage which are ground truth pairs, 52.3\% of them ranks top-1 in the second stage. Moreover, in the re-ranking stage, we have tried to involve only graph features, and the top-1 and top-3 accuracies are only 45.3\% and 63.3\%, respectively (Table 2), showing that the involvement of text information is still important for improving accuracies. All in all, we conclude that the dramatic improvement in top-1 accuracy may results from the architecture of ranker and the information from ChatGPT involved in the re-ranking stage.
\subsection{Generalization to Rare Reaction Types}
To assess the generalizability of the T-Rex model, we computed accuracies across all 10 reaction types in one side of cross-dataset experiments (USPTO-MIT$\rightarrow$USPTO-50k). Notably, the top-1 accuracy across all reaction types reveals that the T-Rex model consistently outperforms G2Gs, demonstrating superior performance. And T-Rex outperforms G2Gs greatly in terms of accuracies for some rare reaction types including Oxidations, Functional group inter-conversion (FGI) as well as Reductions. Moreover, the top-1 accuracies of G2Gs and T-Rex for these specific reaction types within the test dataset along with the corresponding proportions of these types are presented in Table \ref{tab:rare}. T-Rex exhibits substantially higher accuracy than G2Gs for these specific types, showing its potential for generalization to rare reaction types.

\begin{table}[htbp]
\begin{tabular}{lcc}
\hline
\textbf{\begin{tabular}[c]{@{}l@{}}Reaction Type\\ (Proportion in \\ Test Dataset, \%)\end{tabular}} & \multicolumn{1}{l}{\textbf{\begin{tabular}[c]{@{}l@{}}Accuracy of\\ G2Gs (\%)\end{tabular}}} & \multicolumn{1}{l}{\textbf{\begin{tabular}[c]{@{}l@{}}Accuracy of\\ T-Rex (\%)\end{tabular}}} \\ \hline
Oxidations (1.52)                                                                                    & 33.8                                                                                         & 48.6                                                                                          \\
FGI (3.52)                                                                                           & 11.6                                                                                         & 21.5                                                                                          \\
Reductions (9.38)                                                                                    & 3.28                                                                                         & 44.8                                                                                          \\ \hline
\end{tabular}%
\caption{The accuracies for G2Gs and T-Rex for rare reaction types of oxidations, FGI as well as reductions.}
\label{tab:rare}
\end{table}
\begin{figure*}[!t]
\centering
\includegraphics[width=\linewidth]{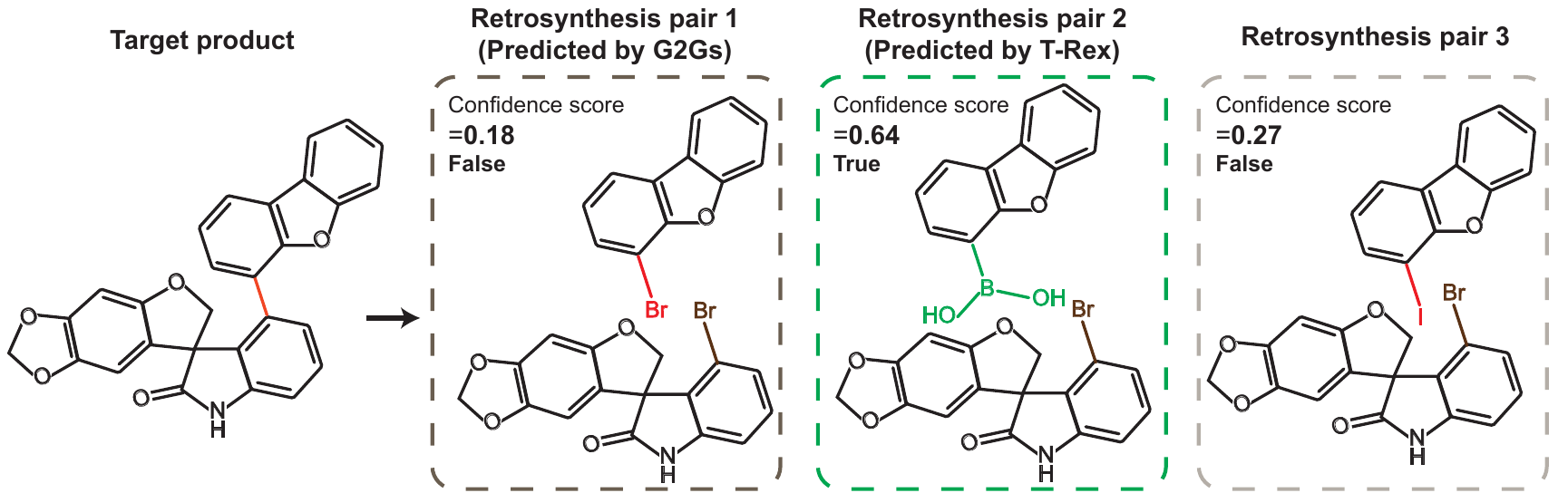}
\caption{\textbf{A case study illustrating how the re-ranking stage can correct the mispredicted reaction center.} The presented reactants are the top 3 predicted retrosynthesis pairs in the first stage. The ground truth is Reactant 2, which is ranked as the second in the first stage but re-ranked as the first in the second stage by our T-Rex model.}
\label{case_study}
\end{figure*}
Moreover, for top-5 and top-10 accuracies, we noticed that G2Gs can outperform T-Rex slightly for some reaction types. For example, for acylation and related processes, taking up 24.3\% of the test dataset, the top-5 and top-10 accuracies for G2Gs are 64.8\%, 70.2\% while the top-5 and top-10 accuracies for T-Rex are 63.0\%, 66.2\%. However, for some rare reaction types, T-Rex instead outperforms G2Gs greatly. For example, for Functional group addition (FGA) reactions(taking up 0.52\% in the test dataset), the top-5 and top-10 accuracies for G2Gs are both 0\% while the top-5 and top-10 accuracies for T-Rex are both 12\%; for FGI reactions (taking up 3.52\% in the test dataset), the top-5 and top-10 accuracies for G2Gs are 26.7\%, 34.3\% while the top-5 and top-10 accuracies for T-Rex are 30.3\%, 40.2\%. From these types, we can still see T-Rex's potential of generalization to rare reaction types. 

\subsection{Case Study}
We presented the t-SNE visualizations of the embeddings of G2Gs, G2Gs with text description(G2Gs+Text), and our T-Rex model in \textbf{Fig.}~\ref{tsne}. Here, G2Gs and G2Gs+Text show the embeddings from the reaction center identification stage, while T-Rex shows the embeddings from the re-ranking stage, which are unavailable for the other two methods. We can see that the embeddings of both G2Gs and G2Gs+Text model do not show visible patterns while our T-Rex can show clear pattern for the retrosynthesis pairs, again demonstrating the effectiveness of the re-ranking stage. Furthermore, we visualized an example that the re-ranking stage successfully correct the mis-predicted reaction center by using the text information  (\textbf{Fig.}~\ref{case_study}). More examples are presented in Section \ref{examples} in the Appendix.

\section{Related Work}
Due to the large search space in retrosynthesis, many machine learning approaches have been developed for retrosynthesis prediction. Existing approaches can be classified into template-based and template-free models. These templates are either hand-crafted by experts~\cite{hartenfeller2011collection, szymkuc2016computer} or extracted from large databases automatically~\cite{coley2017prediction, law2009route}. Prior works have exploited various techniques to select from multiple candidate templates ~\cite{GLN,USPTO-50k, segler2017neural, LocalRetro, zhu2023mathcal, baylon2019enhancing}. Nevertheless, the process of identifying templates is extremely time-consuming. To address this issue, template-free methods~\cite{liu2017retrosynthetic, karpov2019transformer, tu2022permutation, sacha2021molecule, somnath2020learning, Dual-TF, MEGAN} utilize SMILES~\cite{SMILES} representation to formulate the task as a sequence-to-sequence generation problem. G2Gs~\cite{G2Gs} converted SMILES strings to graph structures in order to capture the intrinsically rich chemical contexts. To the best of our knowledge, none of these approaches have incorporated text data, especially those generated by pre-trained language models, into the prediction of retrosynthesis. In contrast, our T-Rex framework exploited text descriptions of molecules as additional features for modeling molecular graphs, achieving substantial improvement on two retrosynthesis prediction datasets.

ChatGPT and similar large models have recently transformed natural language processing~\cite{leiter2023chatgpt,bordt2023chatgpt, desaire2023chatgpt, siu2023chatgpt, uludag2023use, juhi2023capability, bubeck2023sparks}. Due to their success in NLP, there is a growing interest in utilizing ChatGPT to aid scientific research~\cite{auer2023sciqa, castro2023large}, such as drug discovery~\cite{sharma2023chatgpt, liang2023drugchat}, computational materials science~\cite{hong2023chatgpt}, and chemistry~\cite{mahjour2023designing}. Nevertheless, the majority of these approaches primarily focus on text data only, overlooking the possibility of using ChatGPT to advance the modeling of other non-text data modalities in scientific domain. In contrast, we investigated how ChatGPT can improve retrosynthesis prediction through combining the textual description generated by ChatGPT and molecular graphs. 

\section{Conclusion}
In this paper, we have introduced T-Rex, a novel retrosynthesis prediction model utilizing the textual descriptions generated by large langauge models. Specifically, we have integrated the text embeddings generated by ChatGPT into the graph model. We have also introduced the ranking stage to further improve the performance. The experimental results show that the involvement of text information and the ranking stage enables the model to outperform existing template-free models in both datasets. 

To the best of our knowledge, our framework is the first attempt that leverages text description for retrosynthesis prediction. Since large language models are trained on massive-scale corpora, they often possess powerful reasoning and analysis capabilities. We envision that our method will motivate future research in using text language models for other tasks in computational chemistry.

\bibliography{anthology,custom}
\bibliographystyle{acl_natbib}

\appendix
\section{Detailed Results}
We display the detailed results of comparison models on USPTO-50k and filtered USPTO-MIT datasets in \textbf{Tables}~\ref{tab:app1},~\ref{tab:app2} and~\ref{tab:app3}. The results of G2Gs model are from \url{https://torchdrug.ai/docs/benchmark/retrosynthesis.html}, which is slightly different from those in~\cite{G2Gs}.
\label{sec:appendixA}
\begin{table}[h]
\centering
\begin{tabular}{lllll}
\hline
\multicolumn{1}{c}{\multirow{2}{*}{USPTO-50k}} & \multicolumn{4}{c}{Accuracy}                                    \\ \cline{2-5} 
\multicolumn{1}{c}{}                        & Top-1             & Top-3             & Top-5             & Top-10                   \\ \hline
\multicolumn{5}{c}{Template-free} \\ \hline
Transformer                                  & 0.379        & 0.573          & 0.627         & \multicolumn{1}{c}{$\backslash$} \\
RetroXpert & 0.502 & 0.613 & 0.623 & 0.635\\
MEGAN & 0.479 & 0.712 & \textbf{0.785} & \textbf{0.857}\\
G2Gs                                         & 0.437          & 0.676          & 0.769          & 0.831          \\
G2Gs+Text                                         & 0.441          & 0.705          & 0.784          & 0.852          \\
T-Rex                                     & \textbf{0.523} & \textbf{0.735} & \textbf{0.785} & 0.852 \\ \hline
\multicolumn{5}{c}{Template-based} \\ \hline
GraphRetro                                         & 0.537 & 0.683 & 0.722 & 0.755          \\
LocalRetro                                        & 0.534 & 0.775 & 0.859 & 0.924          \\
$\mathcal{O}$-GNN                                         & 0.541 & 0.777 & 0.860 & 0.925          \\\hline
\end{tabular}
\caption{Top-k exact match accuracy for USPTO-50k dataset when reaction class is not given for template-free models including Transformer~\cite{karpov2019transformer}, RetroXpert~\cite{yan2020retroxpert}, MEGAN~\cite{MEGAN}, G2Gs~\cite{G2Gs} as well as T-Rex. We also include some state-of-the-art template-based models including GraphRetro~\cite{somnath2020learning}, LocalRetro~\cite{LocalRetro} and $\mathcal{O}$-GNN~\cite{zhu2023mathcal} for comparison.}
\label{tab:app1}
\end{table}
\begin{table}[h]
\centering
\begin{tabular}{lllll}
\hline
\multicolumn{1}{c}{\multirow{2}{*}{USPTO-MIT}} & \multicolumn{4}{c}{Accuracy}                                    \\ \cline{2-5} 
\multicolumn{1}{c}{}                         & Top-1             & Top-3             & Top-5             & Top-10                     \\ \hline
G2Gs                                      & 0.417          & 0.630          & \textbf{0.698}          & 0.750 \\
G2Gs+Text                                  & 0.420          & 0.631          & 0.697          & \textbf{0.753}                     \\
T-Rex                                     & \textbf{0.538} & \textbf{0.675} & 0.694 & \textbf{0.753}            \\ \hline
\end{tabular}
\caption{Top-k exact match accuracy for USPTO-MIT dataset.}
\label{tab:app2}
\end{table}
\begin{table}[h]
\centering
\begin{tabular}{lllll}
\hline
\multicolumn{1}{c}{\multirow{2}{*}{USPTO-50k}} & \multicolumn{4}{c}{Accuracy}                                    \\ \cline{2-5} 
\multicolumn{1}{c}{}                         & Top-1             & Top-3             & Top-5             & Top-10                   \\ \hline
\multicolumn{5}{c}{Template-free} \\ \hline
Seq2seq                                      & 0.374          & 0.524          & 0.570          & 0.617 \\
MEGAN & 0.616 & 0.831 & 0.881 & 0.925\\
G2Gs                                         & 0.625          & 0.849          & 0.904          & 0.935                     \\
T-Rex                                     & \textbf{0.630} & \textbf{0.858} & \textbf{0.912} & \textbf{0.959}            \\ \hline
\multicolumn{5}{c}{Template-based} \\ \hline
GraphRetro                                         & 0.639 & 0.815 & 0.852 & 0.881          \\
LocalRetro                                         & 0.639 & 0.868 & 0.924 & 0.963          \\
$\mathcal{O}$-GNN                                         & 0.657 & 0.877 & 0.934 & 0.969          \\\hline
\end{tabular}
\caption{Top-k exact match accuracy for USPTO-50k dataset when reaction class is given for template-free models including Seq2Seq~\cite{liu2017retrosynthetic}, MEGAN~\cite{MEGAN}, G2Gs~\cite{G2Gs} as well as T-Rex. Since the performance of G2Gs on the paper and the website are different, and the G2Gs can outperform MEGAN on top-k accuracy for USPTO-50k when reaction class is given, we still assert that G2Gs is one of the state-of-the-art model. We also include some state-of-the-art template-based models including GraphRetro~\cite{somnath2020learning}, LocalRetro~\cite{LocalRetro} and $\mathcal{O}$-GNN~\cite{zhu2023mathcal} for comparison.}
\label{tab:app3}
\end{table}

\section{Discussion about the Potential Data Leakage Issue}
To investigate potential data leakage issues in ChatGPT, we conducted an in-depth analysis of the text generated by the model. In our examination of the generated texts, only 3\% of the texts mention one of the reactant names and only 0.1\% of the texts mention both names. Especially, none of the texts mention the SMILES of any reactant. Furthermore, it was observed that the deletion of these specific items led to a marginal decrease of 0.2\% in top-1 accuracy on the test dataset. Additionally, MolT5, an alternative model for generating molecule descriptions, was employed. The texts generated by MolT5 exhibited an absence of any mention of reactants. The findings presented in Table \ref{tab:ticl} underscore the significance of text information quality. In summary, the improvement of our methodology primarily comes from considering the text information rather than the potential data leakage of ChatGPT.

\section{Stereochemistry Issue}
In many prior retrosynthesis prediction studies, results have commonly been presented under the assumption of stereochemistry ignorance, where all isomers of the ground truth are deemed correct. Consequently, within the main body of this paper, we adhere to the same assumption to ensure a meaningful comparison with earlier works, including G2Gs. However, for a fair assessment of other models such as LocalRetro ~\cite{LocalRetro}, we also showcase results for USPTO-50k with stereochemistry taken into account in Table \ref{tab:stereo}.
\begin{table}[h]
\centering
\begin{tabular}{lllll}
\hline
\multicolumn{1}{c}{\multirow{2}{*}{USPTO-50k}} & \multicolumn{4}{c}{Accuracy}                                    \\ \cline{2-5} 
\multicolumn{1}{c}{}                        & Top-1             & Top-3             & Top-5             & Top-10                   \\ \hline
G2Gs                                         & 0.425          & 0.661          & 0.749          & 0.813          \\
G2Gs+Text                                         & 0.429          & 0.688          & 0.764          & \textbf{0.831}          \\
T-Rex                                     & \textbf{0.510} & \textbf{0.717} & \textbf{0.766} & \textbf{0.831} \\ \hline
\end{tabular}
\caption{Top-k exact match accuracy for USPTO-50k dataset when reaction class is not given.}
\label{tab:stereo}
\end{table}

It can be seen from the table that under the circumstance of stereochemistry, the T-Rex still outperforms the G2Gs model for top-1, 3, 5 and 10.

\section{Limitations of T-Rex}
Our T-Rex model currently has the following limitations. First, the utilization of ChatGPT API and MolT5 model to generate auxiliary text information requires us to run ChatGPT APIs a large number of times, resulting in substantial costs. Meanwhile, it takes 9 hours for MolT5 model to generate the texts for the classifier stage on a single NVIDIA RTX A4000, which can only achieve a moderate accuracy of 70\% for reaction type prediction. Therefore, we plan to exploit active learning to decrease the number of times we need to run large language models. Moreover, our approach can only be applied to one-step retrosynthesis prediction for single products. It would be interesting to extend our T-Rex framework to support multi-step retrosynthesis and explore how ChatGPT can be used in different steps.

\section{Examples of Generated Texts}
\label{examples}In this section, we present several examples of text generated by ChatGPT along with accompanying analyses. Additionally, we juxtapose MolT5-generated texts with those generated by ChatGPT to highlight both the remarkable capabilities and limitations of ChatGPT.
\begin{table*}[p]
\centering
\begin{tabularx}{1.05\textwidth}{p{0.2\textwidth}p{0.8\textwidth}}
\hline
\textbf{Item} & \textbf{Value}             \\ \hline                                              
Product SMILE    & Oc1ccc(OCc2ccccc2)c(Br)c1 \\ \hline
Product IUPAC name   & 3-bromo-4-phenylmethoxyphenol(4-(Benzyloxy)-2-bromophenol) \\ \hline
Reactant smiles     & BrCc1ccccc1+Oc1ccc(O)c(Br)c1 \\ \hline
Reactant IUPAC names  & bromomethylbenzene+ 2-bromobenzene-1,4-diol \\ \hline
MolT5-generated text  & The molecule is a cyclic ketone that is then2(2-bromo)-derivative of cyclopropane. It has a role as a metabolite and a mouse metabolite. It is a cyclic ketone and a spiro compound. It derives from a cyclopropane. 

\textbf{\textcolor{red}{Problem with MolT5-generated text: it is not cyclic ketone and it has nothing to do with spiro compound. Evidently it is not a derivative of cyclopropane, either.}}\\ \hline
ChatGPT-generated text  & 3-Bromo-4-phenylmethoxyphenol is an organic compound with a molecular formula of C10H9BrO. It is a white crystalline solid with a melting point of 212-214°C. Its structure consists of a benzene ring with an -OH group attached to the 4th carbon atom, a bromine atom bonded to the 3rd carbon atom, and a -OCH3 group attached to the 4th carbon atom.

The most likely reaction to synthesize 3-bromo-4-phenylmethoxyphenol would involve the following steps:

\hspace{0.3cm}1. Start with benzene and treat it with a halogenating agent such as bromine to add a bromine atom to the 3rd carbon of the benzene ring. This will form 1,3-dibromobenzene.

\hspace{0.3cm}2. Reduce the 1,3-dibromobenzene with a reducing agent such as zinc to form 3-bromophenol.

\hspace{0.3cm}3. Treat the 3-bromophenol with an alcohol in presence of an acid catalyst such as sulfuric acid to form 3-bromo-4-phenylmethoxyphenol. 

\textbf{\textcolor{brown}{Analysis of the ChatGPT-generated text: ChatGPT identified some of the physical features, but the melting point is wrong(actually it is 72-73°C according to www.chemicalbook.com). Moreover, it analyzed its molecular structure but not exactly right. For predicting the synthetic route, it identified a possible reaction precursor: 3-bromophenol. However, in practice, it is hard to directly add a great organic group to the benzene ring so it does not match the ground truth reaction. Therefore, we think that ChatGPT can provide useful information for local structures, but it is challenging for ChatGPT to identify more detailed information about the product.}}\\ \hline
\end{tabularx}
\caption{Example of 3-bromo-4-phenylmethoxyphenol(4-(Benzyloxy)-2-bromophenol).}
\end{table*}

\begin{table*}[p]
\centering
\begin{tabularx}{1.05\textwidth}{p{0.2\textwidth}p{0.8\textwidth}}
\hline
\textbf{Item} & \textbf{Value}             \\ \hline                                              
Product SMILE    & O=C1CN(c2ccncc2)CCN1c1ccc(OCc2ccccc2)cc1 \\ \hline
Product IUPAC name   & 1-(4-phenylmethoxyphenyl)-4-pyridin-4-ylpiperazin-2-one \\ \hline
Reactant smiles     & Brc1ccc(OCc2ccccc2)cc1+O=C1CN(c2ccncc2)CCN1 \\ \hline
Reactant IUPAC names  & 1-bromo-4-phenylmethoxybenzene+ 4-pyridin-4-ylpiperazin-2-one \\ \hline
MolT5-generated text  & The molecule is a tertiary amino compound that is up toc2c(naCc3cccc3) with an additional cinchoyl group. It has a role as a vasodilator agent, an anti-ulcer drug and a metabolite. It is a tertiary amino compound and an epoxide. It derives from a hydride of a nipecoside.

\textbf{\textcolor{red}{Problem with MolT5-generated text: it is neither tertiary amino nor epoxide. Moreover, it contains some meaningless words including nipecoside and cinchoyl.}}\\ \hline
ChatGPT-generated text  & 1-(4-Phenylmethoxyphenyl)-4-pyridin-4-ylpiperazin-2-one is an organic compound composed of several functional groups. It has a phenyl group attached to a methoxy group, which is attached to a pyridin-4-yl group, which is attached to a piperazin-2-one group.

The most possible reaction to synthesize this compound step by step would be a sequence of nucleophilic substitution and cyclization reactions. Firstly, a nucleophilic substitution of a halogenated phenol with a pyridinium chloride would form a pyridinium salt. This salt can then be reacted with a primary amine in a cyclization reaction to form a pyridine intermediate. Finally, this pyridine intermediate can be reacted with a methoxybenzene in a nucleophilic substitution reaction to form the target compound. 

\textbf{\textcolor{brown}{Analysis of the ChatGPT-generated text: ChatGPT identified its phenylmethoxyphenyl group and pyridin-4-yl group which attach to the piperazin-2-one group. However, it contains minor errors about the group structure. Moreover, it predicted a pyridine intermediate as a potential reaction precursor and it also demonstrated that the final step is a nucleophilic substitution reaction. However, it mistakenly regard methoxybenzene as one of the reactants while 1-bromo-4-phenylmethoxybenzene is the true reactant.}}\\ \hline
\end{tabularx}
\caption{Example of 1-(4-phenylmethoxyphenyl)-4-pyridin-4-ylpiperazin-2-one.}
\end{table*}

\begin{table*}[p]
\centering
\begin{tabularx}{1.05\textwidth}{p{0.2\textwidth}p{0.8\textwidth}}
\hline
\textbf{Item} & \textbf{Value}             \\ \hline                                              
Product SMILE    & Cn1c(-c2cc(C(F)(F)F)cc(C(F)(F)F)c2)nc(-c2ccncc2)c1C(=O)N1CCC(N2CCCC2)CC1 \\ \hline
Product IUPAC name   & [2-[3,5-bis(trifluoromethyl)phenyl]-3-methyl-5-pyridin-4-ylimidazol-4-yl]-(4-pyrrolidin-1-ylpiperidin-1-yl)methanone \\ \hline
Reactant smiles     & Cn1c(-c2cc(C(F)(F)F)cc(C(F)(F)F)c2)nc(I)c1C(=O)N1CCC(N2CCCC2)CC1 + OB(O)c1ccncc1 \\ \hline
Reactant IUPAC names  &  [2-[3,5-bis(trifluoromethyl)phenyl]-5-iodo-3-methylimidazol-4-yl]-(4-pyrrolidin-1-ylpiperidin-1-yl)methanone + pyridin-4-ylboronic acid \\ \hline
MolT5-generated text  & The molecule is a cyclic ketone that is nitrilo substituted at position 2 by a methyl group. It is a cyclic ketone, an organofluorine compound, a pyrrolidinone and a monocarboxylic acid amide. It derives from a nitrilol.

\textbf{\textcolor{red}{Problem with MolT5-generated text: it is not cyclic ketone and it does not have anything to do with nitrilo substitution and nitrilol.}}\\ \hline
ChatGPT-generated text  & The structure of 2-[3,5-bis(trifluoromethyl)phenyl]-3-methyl-5-pyridin-4-ylimidazol-4-yl]-(4-pyrrolidin-1-ylpiperidin-1-yl)methanone is a heterocyclic compound composed of two rings, an imidazole ring and a pyridine ring, both of which contain nitrogen atoms. The two rings are connected by a methylene bridge. The imidazole ring is further substituted with a 3,5-bis(trifluoromethyl)phenyl group and the pyridine ring is substituted with a 4-pyrrolidin-1-ylpiperidin-1-yl group. The most likely reaction to synthesize this compound would be an Ullmann-type coupling reaction. This would involve the reaction of an aryl halide (3,5-bis(trifluoromethyl)phenyl bromide) with a primary alkyl halide (4-pyrrolidin-1-ylpiperidin-1-yl bromide) in the presence of a copper catalyst, giving the alkyl aryl ether as a product. This alkyl aryl ether can then be reacted with an imidazole-4-carbaldehyde to produce the desired 2-[3,5-bis(trifluoromethyl)phenyl]-3-methyl-5-pyridin-4-ylimidazol-4-yl]-(4-pyrrolidin-1-ylpiperidin-1-yl)methanone.

\textbf{\textcolor{brown}{Analysis of the ChatGPT-generated text: ChatGPT identifies two of the heterocycles: an imidazole ring and a pyridine ring. It also points out the accurate positions of 3,5-bis(trifluoromethyl)phenyl group and 4-pyrrolidin-1-ylpiperidin-1-yl group. It demonstrated a possible synthetic route using Ullmann-type coupling reaction. However, this route does not match the ground truth which is a Suzuki reaction. Therefore, we think that ChatGPT can provide useful information for local structures, but cannot identify information related to global structures, such as reaction type and synthetic route.}}\\ \hline
\end{tabularx}
\caption{Example of [2-[3,5-bis(trifluoromethyl)phenyl]-3-methyl-5-pyridin-4-ylimidazol-4-yl]-(4-pyrrolidin-1-ylpiperidin-1-yl)methanone.}
\end{table*}
\end{document}